\documentclass[11pt]{article}

\usepackage[preprint]{acl}

\usepackage{times}
\usepackage{latexsym}

\usepackage[T1]{fontenc}

\usepackage[utf8]{inputenc}

\usepackage{microtype}

\newcommand{\rbar}{\overline{r}}
\usepackage{inconsolata}

\usepackage{graphicx}
\usepackage{amssymb}
\usepackage{amsmath}
\usepackage{booktabs}
\usepackage{bm}
\usepackage{enumitem}
\usepackage{cleveref}
\newcommand{\tht}{\bm{\theta}}
\newcommand{\pp}{\mathbf{p}}
\newcommand{\pbar}{\bar{\mathbf{p}}}
\newcommand{\eps}{\bm{\varepsilon}}
\newcounter{takeaway}

\newcommand{\pcom}{\mathbf{p}_{\mathrm{COM}}}

\title{Center-of-Mass Interpretation of Evolution Strategies}

\author{Akshat Gupta \\
  UC Berkeley\\
  \texttt{akshat.gupta@berkeley.edu}}

\begin{document}
\maketitle
\begin{abstract}
Evolution strategies (ES) update a model by sampling a population of perturbed copies, scoring each one, and moving in the reward-weighted direction of the perturbations. This update is usually presented as an estimate of a smoothed gradient. We present a geometrical interpretation of this update instead. Treat each perturbed model as a particle and its reward as the particle's mass. Before scoring, every particle has the same mass and the population balances at its centroid. Once rewards are assigned, it balances at its reward-weighted center of mass instead. We show that the mean-subtracted ES update is exactly this displacement, scaled by a scalar factor that depends on the mean reward, for any finite population. The physical reading needs nonnegative rewards, which can always be arranged by adding a constant to every reward without changing the update. In the limit of a large population or under antithetic sampling, the centroid sits on the model itself, so ES has an even simpler geometric interpretation: from its current location, the model moves towards the reward-weighted center of mass of its own population. The same picture explains what subtracting the mean reward does: it changes the point from which the displacement is measured, from the model to the sampled centroid. Our contribution is an intuitive, geometric way to visualize an ES update, exact for any finite population.
\end{abstract}

\begin{figure*}[t]
  \centering
  \includegraphics[width=1\textwidth]{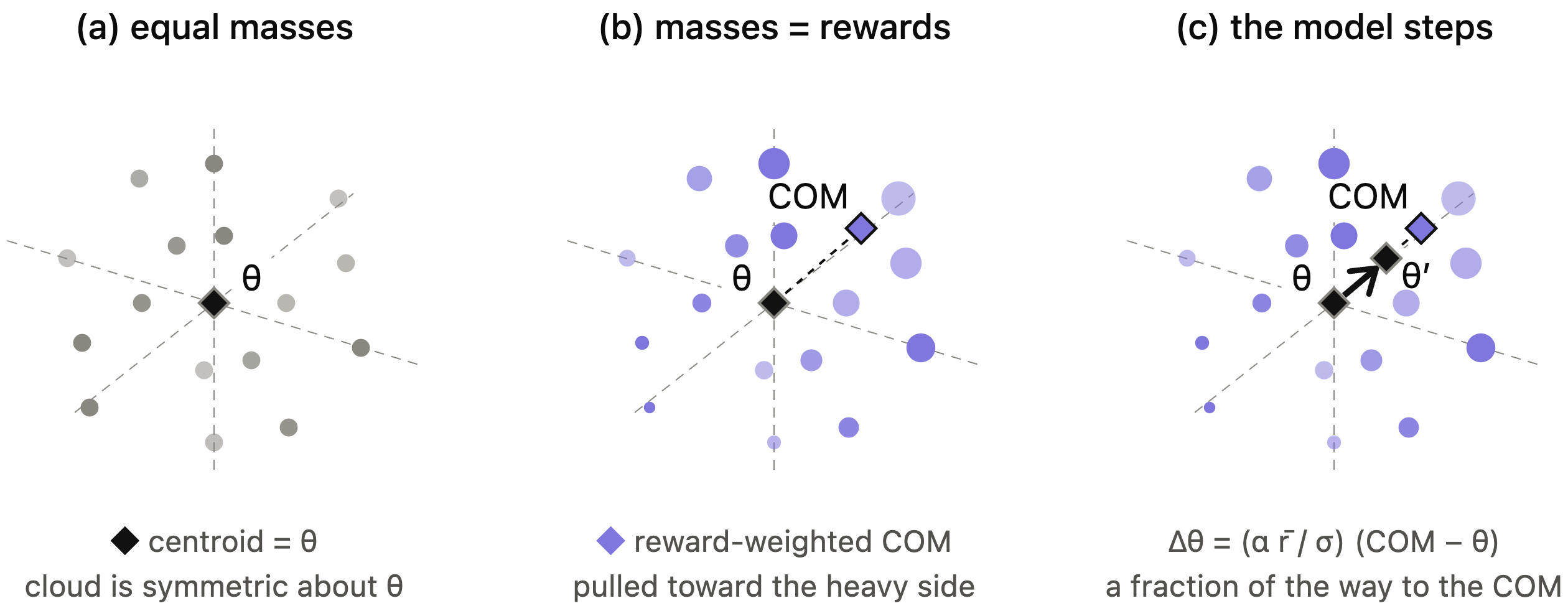}
\caption{One ES update as a change of center of mass.
  (a)~Candidates are sampled around the model $\tht$ in antithetic pairs, so with equal masses the population's centroid coincides with $\tht$.
  (b)~Giving each candidate a mass equal to its reward moves the center of mass (COM) towards the high-reward side.
  (c)~The update is proportional to the displacement from the centroid to the COM, $\Delta\tht=(\alpha\rbar/\sigma)(\pcom-\pbar)$, which here starts at $\tht$. The same sixteen candidates are shown in every panel; displacements are exaggerated for visibility.}
  \label{fig:com}
\end{figure*}

\section{Introduction and Related Work}

Evolution strategies (ES) have recently returned as a practical alternative to backpropagation for fine-tuning large language models \citep{qiu2026es,abdi2026forgetting}. The algorithm is simple. Perturb the current model $N$ times with Gaussian noise, score each perturbed copy, and merge the perturbations weighted by their rewards. The standard way to understand this update is as a Monte Carlo estimate of the gradient of a smoothed objective \citep{nesterov2017random,salimans2017es}. That view is correct, but says little about what a single update does to the model in parameter space.

In this paper, we give the ES update a geometrical interpretation grounded in a physical intuition. Think of the $N$ perturbed models as particles sitting around the current model (Figure \ref{fig:com}(a)). Before scoring, every particle has the same mass and the population balances at its centroid, which is also its center-of-mass (COM). After scoring, give each particle a mass equal to its reward. The COM then shifts away from the centroid. We show that the mean-subtracted ES update is exactly this displacement, scaled by a scalar factor that depends on the mean reward, for any finite population (Figure \ref{fig:com}). For the masses to be physical, the rewards should be nonnegative with a positive total. This is not a restriction: adding a constant to every reward makes it so and leaves the update unchanged. In the limit of a large population or at any $N$ under antithetic sampling (mirrored sampling), the centroid sits on the model itself, and the picture becomes simpler and even more intuitive: \textbf{from its current location in parameter space, the model moves towards the reward-weighted center of mass of its sampled population.}

The center of mass has appeared in the ES literature before, but in a different role. In classical evolution strategies the best offspring are selected and the parent is moved to their centroid, a point that \citet{beyer2002es} describe as a center of mass. CMA-ES does the same with rank-based weights in place of equal ones \citep{hansen2016cma}, and the selection term of Price's equation gives the change in a population's mean trait as the covariance between trait and fitness \citep{price1970selection}. In each case the object described is the new mean, written as a weighted average of the candidates. The mean-subtracted update used in modern practice, for instance by \citet{qiu2026es} with an additional standard-deviation normalization, is not of this form. It takes a step, and we show that this step is the displacement between two centers of mass of the same population, one computed with equal masses and one with rewards as masses. This is also what makes the role of reward centering visible. Subtracting the mean reward changes the point from which the displacement is measured, from the model location to the sampled centroid.

\section{Background: Anatomy of an ES Update}\label{sec:setup}

Let the current model parameters be represented by a point $\theta\in \mathbb{R}^d$ in a $d-$dimensional parameter space. An ES iteration usually consists of three steps: creating perturbed models, scoring them, and combining the sampled perturbations to update the current parameters.

\textbf{1. Perturb.} First, we independently sample $N$ random directions and create perturbed versions of the original model \citep{salimans2017es}. We represent these perturbed versions by $\mathbf{p}_i$, where 
\begin{equation}\label{eq:candidates}
  p_i=\theta+\sigma\epsilon_i,
  \quad \epsilon_i\sim\mathcal{N}(0,I_d),
  \quad i=1,\dots,N.
\end{equation}
Here, $\epsilon_i$ is the sampled direction, $\sigma>0$ controls the scale of the perturbation, $p_i$ is the resulting perturbed model, and $N$ represents the total number of perturbations involved in one ES update step. Geometrically, the points $p_1,\dots,p_N$ can be visualized to form a population of points in parameter space as shown in Figure \ref{fig:com} (a), sampled from a normal distribution centered at $\theta$.

\textbf{2. Score.} Second, we evaluate each perturbed model $p_i$ to obtain a reward $r_i=f(p_i)$, where a higher reward indicates better performance. We denote the mean reward by $\rbar=\frac{1}{N}\sum_{i=1}^N r_i$ and the standard deviation of rewards within the population by $s_r=\sqrt{\frac{1}{N}\sum_{i=1}^N(r_i-\rbar)^2}$.

\vspace{2mm}
\textbf{3. Combine.} Third, we obtain the model update by taking a weighted sum of the sampled directions,
\begin{equation}\label{eq:esgen}
  \Delta\theta=\frac{\alpha}{N}\sum_{i=1}^N w_i\,\epsilon_i,
\end{equation}
where $\alpha>0$ is a learning rate which also absorbs the perturbation scale following \citet{qiu2026es} and each coefficient $w_i$ is
obtained from a transformation of the rewards. The update
$\Delta\theta$ is then added to the current parameters $\theta$. The choice of transformation determines how each candidate contributes to this update.

\paragraph{The Weighting Rules.} ES implementations differ in how they convert rewards into the coefficients $w_i$. Common choices include subtracting the mean reward, standardizing the rewards, and assigning coefficients based on the ranks of the candidates.


Our main derivation covers weighting rules whose coefficients sum to zero, $\sum_{i=1}^N w_i=0.$ Several variants of the ES algorithm that share this condition, as shown in Table \ref{tab:rules}. Rules whose coefficients do not sum to zero, such as using raw rewards directly, are discussed separately in \ref{sec:uncentered}. 

\section{The Sampled Population and Its Centroid}\label{sec:cloud}

The $N$ perturbed models in \cref{eq:pbar} are sampled around the
current model $\tht$, as shown in \cref{fig:com}(a). The centroid of
this sampled population can be written as
\begin{equation}
\label{eq:pbar}
    \pbar
    =
    \frac{1}{N}\sum_{i=1}^{N}\pp_i
    =
    \tht+\sigma\bar\eps,
    \qquad
    \bar\eps
    =
    \frac{1}{N}\sum_{i=1}^{N}\eps_i,
\end{equation}


The centroid has a simple physical interpretation. Suppose every
sampled model is a point particle with the same mass $m$. The center
of mass of the population is
\begin{equation}
\label{eq:equal-com}
    \frac{\sum_{i=1}^{N} m\,\pp_i}{\sum_{i=1}^{N} m}
    =
    \frac{m\sum_{i=1}^{N}\pp_i}{Nm}
    =
    \frac{1}{N}\sum_{i=1}^{N}\pp_i
    =
    \pbar.
\end{equation}
So the centroid of the sampled population is exactly its equal-weight
center of mass.

\paragraph{Where does the centroid lie relative to the model?}
Taking expectations in Equation \ref{eq:pbar} and using
$\mathbb E[\eps_i]=0$,
\begin{equation}
\label{eq:pbar-expectation}
    \mathbb E[\pbar]
    =
    \tht+\frac{\sigma}{N}\sum_{i=1}^{N}\mathbb E[\eps_i]
    =
    \tht,
\end{equation}
so, averaged over repeated populations, the centroid lies exactly at the current model. For one finite population it is
displaced from $\tht$ by $\sigma\bar\eps$. Since the perturbations are
independent with zero mean, the cross terms
$\mathbb E[\eps_i^\top\eps_j]$ vanish for $i\neq j$, and with
$\mathbb E\|\eps_i\|^2=d$,
\begin{align}
    \mathbb E\|\pbar-\tht\|^2
    &=
    \frac{\sigma^2}{N^2}\sum_{i=1}^{N}\mathbb E\|\eps_i\|^2
    =
    \frac{\sigma^2 d}{N}\label{eq:centroid-distance}
\end{align}

As the population grows, the centroid converges to the model,
\begin{equation}
\label{eq:pbar-limit}
    \bar\eps\longrightarrow 0,
    \qquad
    \pbar\longrightarrow\tht
    \qquad
    \text{as } N\to\infty,
\end{equation}
and in this limit the current model is itself the equal-weight center
of mass of the population. If the perturbations are drawn in mirrored pairs $\pm\eps_i$ \citep{salimans2017es}, then $\bar\eps=0$ and $\pbar=\tht$ exactly, even for a finite population.


\section{ES Moves the Model Along a Center-of-Mass Shift}\label{sec:com}

In the previous section we ignored the rewards and gave every perturbed model the same mass. This is the state of the population ``particles'' before they are evaluated. Once each perturbed model $\pp_i$ is scored, the center-of-mass interpretation sets its mass equal to its reward $r_i$, and the question becomes where the population balances now. We derive this in exact form next.

We start with the ES update in \cref{eq:esgen} for the mean-subtracted rule $w_i=r_i-\rbar$. This rule, often with a further rescaling, is the most common choice in practice \citep{wierstra2014nes,mania2018simple,qiu2026es}. We first rewrite the update in terms of the positions $\pp_i$ instead of the perturbations $\eps_i$. Using $\eps_i=(\pp_i-\tht)/\sigma$ from \cref{eq:candidates}, the summation becomes
\begin{align}
  \sum_i w_i\eps_i
  &= \sum_i(r_i-\rbar)\eps_i \nonumber= \frac1\sigma\sum_i(r_i-\rbar)(\pp_i-\tht) \nonumber\\
  &= \frac1\sigma
     \left[
       \sum_i(r_i-\rbar)\pp_i
       -
       \tht\sum_i(r_i-\rbar)
     \right] \nonumber\\
  &= \frac1\sigma\sum_i(r_i-\rbar)\pp_i.
  \label{eq:positions}
\end{align}

We see that the terms containing $\tht$ cancel because $\sum_i(r_i-\rbar)=0$. For the same reason, any fixed point can be subtracted from every $\pp_i$ in \cref{eq:positions} without changing the summation. Substituting \cref{eq:positions} into \cref{eq:esgen} and subtracting the sampled centroid $\pbar$ from \cref{eq:pbar}, we get
\begin{align}\label{eq:centered-position}
  \Delta\tht
  &=
  \frac{\alpha}{N\sigma}
  \sum_i (r_i-\rbar)\pp_i\nonumber\\
  &=
  \frac{\alpha}{N\sigma}
  \sum_i (r_i-\rbar)(\pp_i-\pbar)\nonumber\\
  &=
  \frac{\alpha}{N\sigma}
  \left[
    \sum_i r_i(\pp_i-\pbar)
    -
    \rbar\sum_i(\pp_i-\pbar)
  \right]\nonumber\\
  &=\frac{\alpha}{N\sigma}\sum_i r_i(\pp_i-\pbar)
\end{align}

because $\sum_i(\pp_i-\pbar)=0$. Let the total reward given to all perturbations be $R=\sum_{i=1}^{N} r_i=N\rbar$. Continuing with the physical analogy, $R$ is the total mass of the system, and when $R\neq0$\footnote{Note that the update itself is well defined for any rewards, but the center-of-mass picture needs $r_i\ge0$ with $R>0$: for $R=0$ the point $\pcom$ is undefined, and for negative rewards the masses are not physical. Both cases are handled by simply adding a constant to every reward, which leaves the update unchanged since $\sum_i(\pp_i-\pbar)=0$.}, the reward-weighted COM of the sampled population is
\begin{equation}\label{eq:reward-com}
    \pcom
    =
    \frac{1}{R}
    \sum_{i=1}^{N} r_i\pp_i.
\end{equation}

We now have all the tools to rewrite the ES update in terms of the two centers of mass:
\begin{equation}\label{eq:com-identity}
\begin{aligned}
  \Delta\tht &= \frac{\alpha}{N\sigma}
  \sum_i r_i(\pp_i-\pbar)\\
    &=
    \frac{\alpha}{N\sigma} \left[ \sum_i r_i\pp_i - \pbar\sum_i r_i \right]\\
    &=
    \frac{\alpha}{N\sigma} \left[ R\pcom - R\pbar\right].
\end{aligned}
\end{equation}
Therefore, we get
\begin{equation}\label{eq:com-update}
    \boxed{
    \Delta\tht
    =
    \frac{\alpha R}{N\sigma}
    \left(\pcom-\pbar\right)
    =
    \frac{\alpha\rbar}{\sigma}
    \left(\pcom-\pbar\right).
    }
\end{equation}

This is the central observation of the paper. For mean-subtracted ES, the update step has a simple geometric interpretation. The update vector is the displacement from the equal-weight balance point of the system, which is the centroid $\pbar$, to its reward-weighted COM $\pcom$, scaled by $\alpha\rbar/\sigma$ and applied at the current model $\tht$ (\cref{fig:com}(b,c)). Here, system refers to the sampled population of models around $\tht$. In the infinite population regime, or at any $N$ under antithetic sampling, the centroid coincides with the current location of the model (\cref{eq:pbar-limit}), which means that the update literally moves the model at $\tht$ towards the reward-weighted COM of the population. Put simply, the ES update can be understood with a physical intuition---treat reward as mass, and in the infinite population regime the model moves towards the COM of its own system of particles.


\subsection{Centered and Raw Rewards Use Different Reference Points}
\label{sec:uncentered}

The appearance of $\pbar$ in the center-of-mass update is a direct
consequence of subtracting the mean reward. To see this, consider the
same ES update using the raw rewards directly:

\begin{align}
    &\Delta\tht_{\mathrm{raw}}
    =
    \frac{\alpha}{N}
    \sum_i r_i\eps_i
    =
    \frac{\alpha}{N\sigma}
    \sum_i r_i(\pp_i-\tht)
    \nonumber\\
    &=
    \frac{\alpha}{N\sigma}
    \left[
        \sum_i r_i\pp_i
        -
        \tht\sum_i r_i
    \right]
    \nonumber=
    \frac{\alpha R}{N\sigma}
    \left(
        \pcom-\tht
    \right)
\end{align}


Compare this with the centered update from \cref{eq:com-update}, the two updates use the same reward-weighted COM, but they measure its
displacement from different reference points:
\begin{equation}
\label{eq:raw-centered-comparison}
    \boxed{
    \begin{aligned}
        \text{raw rewards:}
        &\qquad
        \tht
        \longrightarrow
        \pcom,
        \\[3pt]
        \text{centered rewards:}
        &\qquad
        \pbar
        \longrightarrow
        \pcom.
    \end{aligned}
    }
\end{equation}

Therefore, without centering, the update measures the displacement from the
current model $\tht$ to the reward-weighted COM. After subtracting the
mean reward, the reference point moves from $\tht$ to the equal-weight
COM of the realized sample, $\pbar$.

We can also write the raw-reward update in terms of the centered-reward update of \cref{eq:com-update}, which we now denote $\Delta\tht_{\mathrm{centered}}$ as follows:
\begin{equation}
\label{eq:raw-decomposition}
    \Delta\tht_{\mathrm{raw}}
    =
    \Delta\tht_{\mathrm{centered}}
    +
    \frac{\alpha R}{N\sigma}\,(\pbar-\tht).
\end{equation}

This shows that the raw-reward update is the centered-reward update plus an additional term that depends on two things: (1) the total reward obtained by all the perturbations, and (2) the distance between the centroid of the sampled population and the current model, which exists only because the population is finite. Neither of these has anything to do with the differences between rewards, which is the only thing that carries information about which direction is better. This is what mean subtraction does geometrically. It removes the part of the update caused by the sampled centroid not sitting on the model, a part that is dominated by noise.\footnote{The term is not pure noise, since its expectation works out to $1/N$ of the expected raw update, but by \cref{eq:centroid-distance} its typical size is of order $\alpha|\rbar|\sqrt{d/N}$, so for $d\gg N$ it is dominated by sampling noise.} The extra term vanishes as the population grows, and mean subtraction removes it exactly at any $N$. Antithetic sampling removes it too, since then $\pbar=\tht$ and the raw and centered updates coincide.

\section{Conclusion}

In this paper, we gave the ES update a geometrical interpretation grounded in physical intuition. We treat each perturbed model as a particle and its reward as the particle's mass. Before scoring, all particles have the same mass and the center of mass of the system is simply the centroid of the sampled population. After scoring, the reward-weighted COM shifts away from the centroid, and we show that the mean-subtracted ES update is exactly this displacement, scaled by a factor that depends on the mean reward. This holds for any finite population. In the infinite population regime, or under antithetic sampling, the centroid coincides with the model and the interpretation becomes even simpler. From its current location, the model moves towards the reward-weighted COM of its own system of particles. The same picture also explains the role of mean subtraction. 

\section*{Limitations}
This paper presents an interpretation of the ES update and not a new method. The identity in \cref{eq:com-update} is exact and needs no experimental validation. The physical reading of rewards as masses needs nonnegative rewards with a positive total. This is always achievable by adding a constant to every reward, but the identity itself holds without it, so the physical picture is a convenience rather than a requirement. Our analysis of the term removed by mean subtraction assumes independent Gaussian perturbations with a fixed $\sigma$, and does not cover adaptive step sizes or covariance updates such as those in CMA-ES. Finally, the claim that this term is dominated by noise is made for $d\gg N$, which is the regime of LLM fine-tuning, and may not hold for small problems.



\bibliography{custom}

\appendix

\section{Appendix}

\begin{table*}[ht]
\centering
\small
\renewcommand{\arraystretch}{1.25}
\begin{tabular}{lll}
\toprule
Weighting rule & Coefficient $w_i$ & Used in \tabularnewline
\midrule
Mean-subtracted reward
  & $r_i-\rbar$
  & NES fitness baseline \citep{wierstra2014nes} \tabularnewline
Standardized reward
  & $(r_i-\rbar)/s_r$
  & ARS \citep{mania2018simple}; LLM fine-tuning \citep{qiu2026es} \tabularnewline
Centered rank transform
  & $\dfrac{q_i-1}{N-1}-\dfrac12$
  & OpenAI-ES \citep{salimans2017es,chrabaszcz2018back} \tabularnewline
Rank utilities
  & $u_{q_i}-1/N$
  & NES \citep{wierstra2014nes} \tabularnewline
\bottomrule
\end{tabular}
\caption{Common ES weighting rules. Under the conventions stated
in the text, each rule produces coefficients that sum to zero.
The centered-rank formula is the one in the released OpenAI-ES code;
the paper itself specifies only ``a rank transformation''
\citep{salimans2017es}. CMA-ES uses the positive, truncated analogue
of the rank utilities, $w_i\propto\log(\mu+\tfrac12)-\log i$, which
sums to one rather than zero \citep{hansen2016cma}.}
\label{tab:rules}
\end{table*}



\end{document}